\documentclass[numbered]{trbunofficial}
\usepackage{graphicx}
\usepackage{subfigure}
\usepackage{amsmath}
\usepackage{url}
\usepackage{color}

\usepackage[hidelinks]{hyperref}

\AuthorHeaders{Sun, Feng, Yan, and Liu }
\title{C\lowercase{orner} C\lowercase{ase} G\lowercase{eneration} \lowercase{and} A\lowercase{nalysis} \lowercase{for} S\lowercase{afety} A\lowercase{ssessment} \lowercase{of} A\lowercase{utonomous} V\lowercase{ehicles}}

\author{%
  \textbf{Haowei Sun}\\
  Department of Civil and Environmental Engineering\\
  University of Michigan, Ann Arbor, MI, USA\\
  Email: haoweis@umich.edu\\
  \hfill\break
  \textbf{Shuo Feng, Ph.D., Corresponding Author}\\
  Department of Civil and Environmental Engineering\\
  University of Michigan, Ann Arbor, MI, USA\\
  Email: fshuo@umich.edu\\
  \hfill\break
  \textbf{Xintao Yan}\\
  Department of Civil and Environmental Engineering\\
  University of Michigan, Ann Arbor, MI, USA\\
  Email: xintaoy@umich.edu\\
  \hfill\break
  \textbf{Henry X. Liu, Ph.D.}\\
   Department of Civil and Environmental Engineering\\
   University of Michigan Transportation Research Institute\\
   University of Michigan, Ann Arbor, MI, USA\\
   Email: henryliu@umich.edu\\
 \hfill\break
}

\begin{document}
\maketitle
\section{Abstract}

Testing and evaluation is a crucial step in the development and deployment of Connected and Automated Vehicles (CAVs). To comprehensively evaluate the  performance of CAVs, it is of necessity to test the CAVs in safety-critical scenarios, which rarely happen in naturalistic driving environment. Therefore, how to purposely and systematically generate these corner cases becomes an important problem. Most existing studies focus on generating adversarial examples for perception systems of CAVs, whereas limited efforts have been put on the decision-making systems, which is the highlight of this paper. As the CAVs need to interact with numerous background vehicles (BVs) for a long duration, variables that define the corner cases are usually high dimensional, which makes the generation a challenging problem. In this paper, a unified framework is proposed to generate corner cases for the decision-making systems. To address the challenge brought by high dimensionality, the driving environment is formulated based on Markov Decision Process, and the deep reinforcement learning techniques are applied to learn the behavior policy of BVs. With the learned policy, BVs will behave and interact with the CAVs more aggressively, resulting in more corner cases. To further analyze the generated corner cases, the techniques of feature extraction and clustering are utilized. By selecting representative cases of each cluster and outliers, the valuable corner cases can be identified from all generated corner cases. Simulation results of a highway driving environment show that the proposed methods can effectively generate and identify the valuable corner cases.

\hfill\break%
\noindent\textit{Keywords}: Connected and Automated Vehicles, Corner Case Generation, Safety, Deep Reinforcement Learning
\newpage

\section{Introduction}
   
Testing and evaluation of Connected and Automated Vehicles (CAVs) have been studied for years \cite{kalra2016driving, li2016intelligence, li2018artificial, koren2018adaptive, cui2018development, thorn2018framework, li2019parallel, wang2019development, liu2019road, liu2019safe, li2020theoretical}. To comprehensively evaluate the performance of CAVs, it is crucial to test the CAVs in different scenarios, especially the safety-critical ones. In the naturalistic driving environment (NDE), however, safety-critical scenarios rarely happen, so it is very time-consuming and inefficient to collect corner cases from either on-road test or simulation test of NDE \cite{feng2021NADE, yan2021distributionally}. Therefore, how to purposely and systematically generate corner cases becomes an important problem.

Most existing studies for corner case generation focus on the perception systems of CAVs. Utilizing the methods from the domain of computer vision, researchers aim to generate adversarial examples, which can fool the perception system of CAVs, such as misleading the object classification results \cite{chen2018robust, evtimov2017robust, eykholt2018robust} and hiding pedestrians from the perception results \cite{xie2017adversarial}. Consequently, disturbed CAVs may miss safety-critical information and encounter dangerous situations.

The decision-making system is also essential in keeping the safety driving of CAVs. Even though the perception system is perfect in the sense that every object of interest can be correctly observed and recognized, the failure of decision-making system can still cause severe accidents. However, how to generate corner cases for decision-making systems still lacks investigation. Towards solving this issue, several methods have been recently proposed to generate corner cases of simple scenarios (e.g., cross-walking) using adversarial machine learning techniques \cite{ding2020learning}. For real-world traffic environment (e.g., highway driving environment), however, CAVs need to interact with multiple background vehicles (BVs) for a long duration, so the variables that define the corner cases will be high dimensional. To the best of our knowledge, no existing method can handle such a high dimensionality for corner case generation.

In this paper, a unified framework is proposed for high dimensional corner case generation problem. To address the challenge brought by the high dimensionality, Markov Decision Process (MDP) is introduced to formulate the traffic environment, which can simplify the temporal dependency between different snapshots of the simulation. Compared with existing driving models commonly used in prevailed simulation platforms such as IDM \cite{treiber2000congested}, MOBIL \cite{treiber2016mobil}, and Krauss model \cite{krauss1997metastable}, the MDP-based driving model can incorporate the behavior randomness of the real-world datasets, which makes the simulation more realistic and reliable. Moreover, with the learning results of the RL, the obtained simulation can generate much more corner cases purposely, comparing with the naturalistic driving environment. To further simplify the spatial dependency, the BVs of the environment are assumed to make decisions simultaneously and independently during each time step, which is commonly accepted in existing studies \cite{bareiss2015generalized, weng2020model}. Using naturalistic driving data, the empirical distributions of BV's actions can be obtained for every state. By sampling actions of BVs from the distributions at each time step, the naturalistic driving environment can be essentially generated \cite{feng2021NADE, yan2021distributionally}. Based on this formulation, the corner case generation problem of the driving environment is equivalent to optimizing the behavior policy of BVs to improve the probability of corner cases. To achieve this optimization objective, deep reinforcement learning (DRL) \cite{mnih2015human} techniques are utilized to learn the optimal behavior policy of BVs. With the learned policy, the BVs will behave more aggressively when interacting with the CAVs, and therefore systematically generate the corner cases for the complex driving environment, such as highway driving environment.

Due to the diversity of corner cases, the generated cases can usually be divided into several clusters as well as outliers that do not belong to any specific cluster. Therefore, typical corner cases from each cluster and the outliers can represent the internal property of the generated corner cases, which are referred as "valuable" corner cases hereafter. Given the generated corner cases, it is crucial to identify the valuable corner cases, which are usually more valuable for the evaluation and development of CAVs. To achieve this goal, feature extraction and clustering techniques are introduced. Since the generated corner cases could be high dimensional, Principal Component Analysis (PCA) \cite{wold1987principal} is utilized to reduce the dimension of corner cases and extract the principal features. For the extracted features, clustering methods are applied to identify different clusters of corner cases as well as the specific outliers. Consequently, valuable corner cases can be identified.

To validate the proposed method, experiments in a highway driving environment are simulated. To generate the corner cases, one of the commonly used DRL methods, the dueling deep Q network \cite{mnih2015human}, is applied to learn the optimal behavior policy of BVs. After the feature extraction by PCA, K-means \cite{alsabti1997efficient} and the "density-based spatial clustering of applications with noise" (DBSCAN) algorithms \cite{ester1996density} are utilized to analyze the corner cases. The experiment result shows that the proposed method can effectively generate and analyze valuable corner cases.

The rest of the paper is organized as follows:
First, the related works about corner case generation are introduced. Second, we propose a new unified framework for corner case generation using MDP and DRL techniques. Third, feature extraction and clustering techniques are applied for the corner case analysis. After that, two case studies are provided to validate the proposed corner case generation and analysis methods. Finally, we conclude and discuss future works.

\section{Related Works}
In the CAV testing and evaluation domain, the ability to test the performance of CAV under different scenarios is crucial. Therefore, researchers have been deploying on this field to find efficient and reasonable methods of generating corner cases for both CAV perception testing and decision-making testing area.

\subsection{Corner Case Generation for Vehicle Perception}

In the CAV perception field, one popular research area of generating corner cases is the adversarial examples generation method. Many autonomous vehicle manufacturers such as Tesla, Waymo, etc. have been leveraging neuron networks for perception purposes. However, with the rapid development of deep learning and neuron network training, many researchers have found that with a small perturbation on the training examples, the machine learning system can be fooled \cite{goodfellow2014explaining, yuan2019adversarial, buckner2020understanding}. These contaminated examples are defined as adversarial examples. For example, \citep{szegedy2013intriguing} proposed an gradient-based optimization adversary examples generation method. By minimizing the difference between adversarial examples and normal training examples while modifying the predicted label from the machine learning system, the proposed algorithm can automatically generate adversarial examples for specific neuron networks.

In terms of autonomous vehicle testing, there are also many approaches to interfering with vehicle perception, where most attention is focused on attacking the object detection system. \citep{xie2017adversarial} proposed an attack algorithm of object recognition system by removing the pedestrian segmentation. \citep{kurakin2016adversarial} showed that the perturbation in the physical world instead of the image perception area can lead to fatal errors of the CAV perception system.  Furthermore, \citep{evtimov2017robust} and \citep{eykholt2018robust} designed a real-world stop sign with domain knowledge from adversarial examples and successfully let the object recognition system classify it as a speed limit sign. Several works such as Lu et al. \cite{lu2017no, lu2017standard} showed that the adversarial stop signs in the physical world will not fool the modern CAV perception system as the CAV is continuously moving and detecting the object at each time step. However, \citep{chen2018robust} showed that the perception system of CAV is still vulnerable to specific adversarial examples.

\subsection{Corner Cases Generation for Vehicle Decision-making}

From another aspect, researchers have also been focusing on generating corner cases for CAV decision systems. \citep{ma1999worst} have proposed a worst case evaluation method, which formulated the disturbance generator and the controller as two players in a game, and generated corner cases by finding the worst inputs from steering controllers and integrated chassis controllers. However, the proposed method focuses on a single vehicle, without considering the influence of other traffic participants, which is crucial in the evaluating and testing process of CAV.

To generate cases with multiple traffic participants, researchers introduced the risky index and the probabilistic model of the environment to help generate critical cases. For example, Zhao et al. \cite{zhao2016accelerated, zhao2017accelerated} introduced importance sampling techniques and generated testing cases for car-following and lane-changing maneuvers. To reduce the overvalue problem of worst cases, Feng et al. \cite{feng2020testing, feng2020testing2, feng2020adaptive, feng2020safety,feng2021NADE} defined the maneuver challenge and exposure frequency terms and generated cases on various environment settings, including cut-in scenarios, car-following scenarios, and highway-driving environment. \citep{akagi2019risk} use self-defined risky index and naturalistic driving data to sample critical cut-in scenarios.  \citep{o2018scalable} utilized neuron network and imitation learning to calibrate naturalistic driving model from the NGSIM data, and then a highway with 6 vehicles are chosen to generate testing cases. These critical case generation methods consider both the risky index and the naturalistic probability in the generating process. Although the critical cases are significant for the systematic evaluation of CAVs, the corner cases are also important especially for the vulnerability identification of CAVs, which is complementary to the critical cases. How to generate and identify corner cases with the high coverage, variability, and representativeness remains an open question.

To deal with corner cases with long time duration, researchers introduced markov decision process (MDP) and reinforcement learning (RL) techniques to reduce the temporal complexity. \citep{ding2020learning} modeled the environment as the combination of "blocks" and uses REINFORCE algorithm to generate corner cases. However, the modeling method can only be used in simplified environment and cannot deal with large number of traffic participants. \citep{koren2018adaptive} proposed the Adaptive Stress Testing (AST) method, which introduced monte carlo tree search (MCTS) and deep reinforcement learning (DRL) to solve the pedestrian-crossing problem. In this study, however, it also only involves one autonomous vehicle and one or two pedestrians. \citep{karunakaran2020efficient} utilized the deep q network (DQN) to generate corner cases involving one pedestrian and one autonomous vehicle. However, the action chosen for the pedestrian is very simple, and the risk estimation only concentrates on the responsibility-sensitive safety (RSS) metric \cite{shalev2017formal}, which may become misleading in predicting the crash probability.

To address the limitations of the existing decision-making corner case generation method, this paper proposes a corner case generation method over the decision domain. In most scenarios, the corner case is a sequence of snapshots of the environment with multiple traffic participants. Therefore, the corner cases in real-world traffic environments always have high dimensions. However, existing decision-making corner case generation methods are only validated under simplified scenarios and cannot process highly complex environments. In this paper, we propose a corner case generation method for high dimensional and complex traffic simulations. By modeling the environment with Markov Decision Process (MDP), the complexity of the temporal domain is simplified. Moreover, we model the scenario as interactions between multiple traffic participants, which can handle the curse of dimensionality in space.

\section{Corner Case Generation}

In this section, we propose a unified framework for corner case generation based on MDP formulation and DRL techniques. To address the challenge brought by high dimensionality, we formulate the traffic simulation environment as an MDP. By utilizing the naturalistic driving data, we build the naturalistic driving models (e.g., car-following and lane-changing models). In this way, the naturalistic driving environment can be modeled as the interactions between multiple background vehicles with the naturalistic driving models. To purposely generate the corner cases, we formulate the generation problem as an optimization problem of the driving models of BVs. The goal of the optimization problem is to improve the probability of crashes. By utilizing the DRL techniques, the optimization problem can be solved by learning the behavior policy of BVs. The learned behavior policy essentially leads to aggressive driving models of BVs, which can generate corner cases for the CAV under test. 

\subsection{Problem formulation}

In this paper, the problem formulation is consistent with \cite{feng2020testing, feng2020testing2, feng2020adaptive, feng2020safety, feng2021NADE}. Let $\theta$ describe the pre-determined parameters of the operational design domains (ODD), such as the number of lanes, weather, etc. The definitions of scenario and scene are adopted from \cite{ulbrich2015defining}. Under specific ODD parameters, a scene describes the snapshot of the traffic environment, which includes the states of static elements and dynamic traffic participants (e.g., pedestrians and vehicles). A scenario describes the temporal development among a sequence of scenes. Let $X$ represent the decision variables of the scenario and $s$ denote the state of the scene. In this paper, we only consider background vehicles, and each vehicle has three parameters to describe the overall state: position $p$, velocity $v$, and heading angle $\alpha$. Therefore, the decision variable of one vehicle can be defined as $x = \{p, v, \alpha\}$. In each time step, the vehicle numbers in the CAV's neighborhood are different, so we define the number of observed vehicles in time step $t$ as $m_t$. Then, the scene of time step $t$ can be defined as $s_t = \{x_1^{(t)}, x_2^{(t)}, \cdots, x_{m_t}^{(t)}\}$, and the scenario can be defined as:
\begin{eqnarray}
    X = \{s_0, s_1, \cdots, s_n\}.
\end{eqnarray}

To simplify the problem, we model the traffic environment as a Markov Decision Process (MDP). Given a specific state $s_t$, the "agent" can represent the background vehicles in the environment and make a decision $u_t$ (e.g., accelerations). As the $u_t$ is only determined by state $s_t$, it can also be written as $u_t(s_t)$. Therefore, the scenario can be rewritten as follows:
\begin{eqnarray}
    X=\{s_0,u_0,s_1,u_1,s_2,\cdots,s_n\}\label{scenario_def}.
\end{eqnarray}

Then, we define A as the event of interest (e.g., crash event). For a given scenario $X$, we can clearly identify whether it is the event of interest. Therefore, $P(A|X)$ is known for given $X$. Under specific ODD, the probability of the event of interest can be written as follows:
\begin{eqnarray}
\label{PCAV}
P(A|\theta)=\sum_{X \in D} P(A|X)P(X|\theta),
\end{eqnarray}
where $P(X|\theta)$ denotes the probability of specific scenario $X$ given ODD parameter $\theta$, and $D$ represents the set of available scenarios. By using the notation from Equation \eqref{scenario_def}, we can further decompose $P(X|\theta)$ in a factorized way as:
\begin{eqnarray}
    P(X|\theta)=P(s_0|\theta)\prod_{k=0}^{m-1}{P(s_{k+1}|u_k,s_k,\theta)P(u_k|s_k,\theta)}. \label{MDP_for}
\end{eqnarray}  
Here $P(s_{k+1}|u_k,s_k,\theta)$ is the state transition probability, which means the probability of the occurrence of $s_{k+1}$ given state $s_k$ and action $u_k$, and $P(u_k|s_k,\theta)$ denotes the probability of choosing action $u_k$ at state $s_k$.

In Equation \eqref{MDP_for}, $P(s_0|\theta)$ and $P(s_{k+1}|u_k,s_k,\theta)$ are determined by the environment, and $P(A|X)$ is determined by the CAV under test, which can not be modified. Therefore, to improve the exposure frequency of corner cases ($P(A|\theta)$ in Equation \eqref{PCAV}), we should modify the $P(u_k|s_k, \theta)$ such that the $P(X|\theta)$ can be increased for $X$ where $P(A|X) = 1$. $P(u_k|s_k, \theta)$ denotes the probability of agent choosing action $u_k$ under state $s_k$, so it can be viewed as a stochastic policy. If the policy can be optimized to improve $P(A|\theta)$, the corner cases can be generated more purposely. To achieve this goal, the DRL techniques are utilized to train a new policy as the replacement of $P(u_k|s_k, \theta)$, which will be elaborated as follows.

\subsection{Deep Reinforcement Learning Based Method} 
By replacing $P(u_k|s_k,\theta)$ with the modified behavior policy $\pi_{\rho}(u_k|s_k,\theta)$, we can obtain higher probability of the event of interest. The new definition is shown in the following equation:
\begin{eqnarray}
    P(X|\theta,\rho) = P(s_0|\theta) \prod_{k=0}^{m-1}{P(s_{k+1}|u_k,s_k,\theta)\pi_{\rho}(u_k|s_k,\theta)}.
\end{eqnarray}
The newly defined $\pi_{\rho}(u_k|s_k,\theta)$ represents a behavior policy to be learned in DRL problem. Detailed optimization formulation can be seen in Equation \ref{optimization}:
\begin{eqnarray}
\max_{\rho} \quad & P(A|\theta,\rho) = \sum_{X\in D} P(A|X) P(X|\theta,\rho)  .\label{optimization}
\end{eqnarray}

The optimization problem can be considered as training a specific behavior policy in the simulation environment, which will let BVs aggressively interact with the CAV. Therefore, given the complexity of the environment, DRL techniques can have a good performance in solving the optimization problem. As an unsupervised algorithm, DRL techniques can learn optimal policy from experience given specific reward settings. By implementing the Reinforcement Learning (RL) algorithms and using Deep Neuron Network (DNN) as the function approximator, DRL can solve the problem with high complexity and derive well-behaved policy in aspects of video games \cite{mnih2015human}, robotics \cite{tai2017virtual}, etc. Following the ideas of traditional RL, DRL also has three different approaches: value-based DRL, policy-based DRL, and actor-critic DRL \cite{sutton2018reinforcement, mnih2013playing, van2015deep, mnih2015human, hessel2017rainbow, wang2016dueling, lillicrap2015continuous}. In this paper, we mainly use the value-based DRL to solve the optimization problem.

Value-based DRL, also commonly known as the Deep Q Network (DQN), aims at learning the optimal state-action value function $Q(s_t, u_t)$ by estimating the expected reward of action $u_t$ given state $s_t$. Researchers introduce DNN here to represent the $Q$ function, where the neuron network will accept the state and action information as the input and return the estimated state-action value. Specifically, DQN uses a neuron network with parameter $\rho$ to approximate the optimal state-action value function:
\begin{eqnarray}
Q^*_{\rho}(s, u) = \max_{\pi} E[\sum_{i=0}^{\infty}\gamma^i r_{t+1}| s_t=s, u_t=u, \pi_{\rho}],
\end{eqnarray}
where $r_t$ denotes the immediate reward received at step $t$. Therefore, with the optimal state-action value function, we can easily derive the optimal policy which can increase the probability of corner cases. The optimal policy will be a deterministic policy as follows:
\begin{eqnarray}
P(u_t|s_t)=
\begin{cases}
1& u_t=\arg\max_{u}{Q(s_t,u)}\\
0& \text{otherwise}
\end{cases}.
\end{eqnarray}
By only giving crash events positive reward $r_A$, with the training process of the DQN agent, the optimal value function and the optimal policy will bring about higher probability of crash events.

\begin{figure}[h!]
  \centering
  \includegraphics[width=0.6\textwidth]{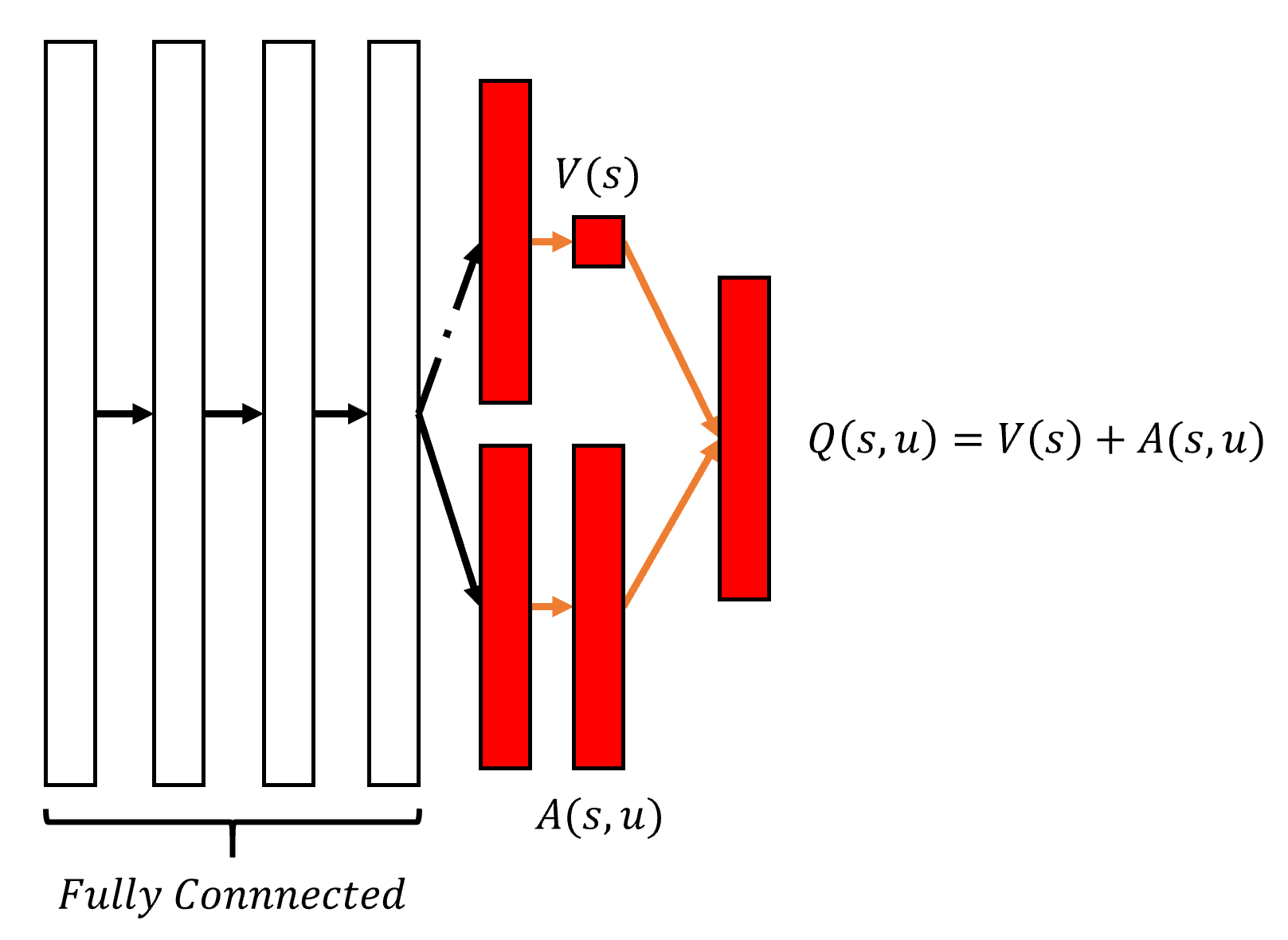}
  \caption{Dueling Network Architecture}\label{fig:dueling}
\end{figure}

In order to further improve the performance of DRL techniques, dueling network architecture \cite{wang2016dueling} are introduced. By defining the new definition value function $V(s)$ and advantage function $A(s,u)$, the dueling network architecture can estimate the value function of the state-action more precisely. The connection between dueling network and the DQN can be seen as follows:
\begin{eqnarray}
A(s,u) = Q(s,u) - V(s). \label{dueling_def}
\end{eqnarray}
Therefore, two neuron networks are introduced to represent $A(s,u)$ and $V(s)$. These two networks often shares some hidden layers as shown in Figure \ref{fig:dueling}.

\section{Corner Case Analysis}

With the learned behavior policy, a large number of corner cases will be generated by simulating the interaction between the agent and the environment. With a detailed understanding of generated corner cases, the evaluation and development of CAVs can be more targeted and effective.

Due to the diversity of corner cases, the generated cases can usually be divided into several clusters as well as outliers that do not belong to any specific cluster. Therefore, typical corner cases from each cluster and the outliers can represent the internal property of the generated corner cases, which are referred as valuable corner cases. To identify the valuable corner cases, we need two following techniques: feature extraction and clustering. The long period and a large number of traffic participants lead to the high dimension of the corner cases, which further bring about difficulty in the analysis. To address the difficulty, we introduce the feature extraction method to reduce the dimension and extract key information from the cases. In this paper, we utilize the Principle Component Analysis (PCA) algorithm to reduce the dimension. Moreover, corner cases can usually be classified into different clusters due to the diversity of the generated corner cases. Different types of cases may differ significantly in terms of frequency. Some types may occupy the majority of corner cases while other types may become the minority. To keep the diversity, the majority and minority need to be analyzed separately. To achieve this goal, the DBSCAN method is used to separate the majority and minority of corner cases as two groups. Then, for each group of the corner cases, the K-means method is utilized to cluster the corner cases, respectively. As unsupervised methods, the DBSCAN and K-means methods can be applied to cluster generic corner cases without knowing the predefined labels.

\subsection{Corner Case Feature Extraction}

Corner cases are sequences of continuous snapshots of traffic environment, which always contain a large number of traffic participants. Therefore, corner cases always have high dimensions, which will bring about difficulties in the analysis. Feature extraction techniques can be utilized to reduce the dimension of the feature space while keeping essential information. Principal component analysis (PCA) \cite{wold1987principal} is one of the most popular feature extraction methods. By maximizing the variance in every direction, the PCA method can project data points from high dimensional space to low dimensional space and transform the data points to a new coordinate system. The first $k$ directions given by PCA are also the directions on which the largest variance of data projection lies. Therefore, in most cases, only picking several coordinates given by PCA will bring about essential information of the original data. By introducing the PCA algorithm, we can extract several key features and simplify the analysis process in the next steps. Additionally, PCA can differentiate and visualize the distance and relatedness between different populations (i.e., data clusters). Therefore, the data points after the PCA projection will be more suitable for the next-step clustering analysis.

\subsection{Corner Case Clustering}

In the generated corner cases, some normal scenarios may occupy a large proportion. For example, some libraries may contain a large number of homogeneous cases (e.g., many rear-end collisions). However, the value of normal scenarios is limited, as the increase in normal corner cases will not bring about much new information for CAV performance. Instead, we should pay more attention to valuable scenarios in the generated corner cases, such as typical corner cases of a specific corner case cluster, and the outliers that have significantly different properties compared with the majority of corner cases. In this paper, to detect the outliers, we utilize the density-based spatial clustering of applications with noise (DBSCAN) method \cite{ester1996density}, which groups the data points that are closely packed together. Therefore, the outliers can be identified as the data points with the low density, namely, the minority of corner cases.

After differentiating the majority and the minority, further analysis can be applied separately. Due to the diversity of the generated corner cases, different cases have different internal patterns and are always distributed in different areas of the feature space. Therefore, by applying the clustering method, corner cases can be classified as different types. Additionally, some typical cases can represent a large number of corner cases (in the same cluster). To automatically classify the type of different data points, the clustering method is commonly used. In this domain, the K-means method \cite{alsabti1997efficient} is among the most popular algorithms. As a non-parametric unsupervised learning method, K-means provides the clustering result minimizing the in-group variance (squared Euclidean distances) for a given objective cluster number, which can help differentiate clusters with different internal properties.

\subsection{Valuable Corner Case Identification}

After corner case clustering, valuable corner case extraction becomes the next topic. The value of one specific corner case represents how much it can help in evaluating CAVs and improving the performance of CAVs. From this perspective, two different types of cases can be defined as valuable: typical cases and rare cases. Typical cases usually represent a large number of corner cases from the same cluster, and rare cases are usually the outliers, which have distinct properties compared with the majority and rarely happen in the NDE. Correspondingly, two techniques are utilized to extract valuable corner cases. First, after data clustering of corner cases, the typical corner cases can be selected by the distance to the center of a specific cluster. Second, by outlier detection methods powered by clustering, rare corner cases with various properties can be extracted. Then, the identification of valuable corner cases can be achieved. In this way, the identified valuable corner cases can well balance the consideration of coverage, variability, and representativeness of the cases. 

\section{Case Study}

In this section, we validate our methodology using experiments in the highway simulation platform. This section consists of four parts. First, we introduce the naturalistic driving data (NDD) processing and NDD driving models. Second, we introduce the simulation platform used in the case study. Third, the proposed corner case generation method is validated in the simulation platform. Finally, by utilizing the corner case analysis method, we extract the valuable corner cases from different experiment settings.

\subsection{Naturalistic Driving Data Processing}

To build the naturalistic highway driving model, we implement a data-driven stochastic model from the Integrated Vehicle-Based Safety Systems (IVBSS) dataset \cite{ference2006integrated} at the University of Michigan, Ann Arbor. By integrating the forward collision warning (FCW), lane departure warning (LDW), lane change warning (LCW), and curve speed warning (CSV) function on passenger cars and heavy trucks, the project aims to prevent rear-end and other crashes. This project collected 650,000 miles of driving data on heavy trucks and 175,000 miles of driving data on regular vehicles.

To calibrate the data-driven naturalistic driving model for the highway environment, we collected data points in which the velocity of vehicles is between $20m/s$ and $40m/s$. As a result, we collected around $3\times 10^{6}$ data points and built up the empirical distribution of BV's action for every state. By sampling actions from the empirical distribution, all BVs are essentially controlled by the naturalistic driving model, which formulate the naturalistic driving environment (NDE).

\subsection{Simulation Platform}

Our simulation platform is based on the open-source simulation platform HIGHWAY-ENV \cite{highway-env}. This simulation environment is fully compatible with OpenAI gym environment \cite{gym2016}, so it can be used to train the autonomous vehicle planning algorithm. In the HIGHWAY-ENV environment, the IDM car-following model \cite{treiber2000congested} and MOBIL \cite{treiber2016mobil} lane-changing model are used to provide continuous traffic flow and reasonable vehicle behaviors. However, default vehicle models are deterministic and can not perform the naturalistic nature of vehicle behavior, which are not suitable for the corner case generation process. To overcome this limitation, we re-design the simulation environment and add control API to improve the controllability of BVs, so the BVs can be controlled by the naturalistic driving model.

\subsection{Corner Case Generation}

\begin{figure}[h!]
  \centering
  \includegraphics[width=0.7\textwidth]{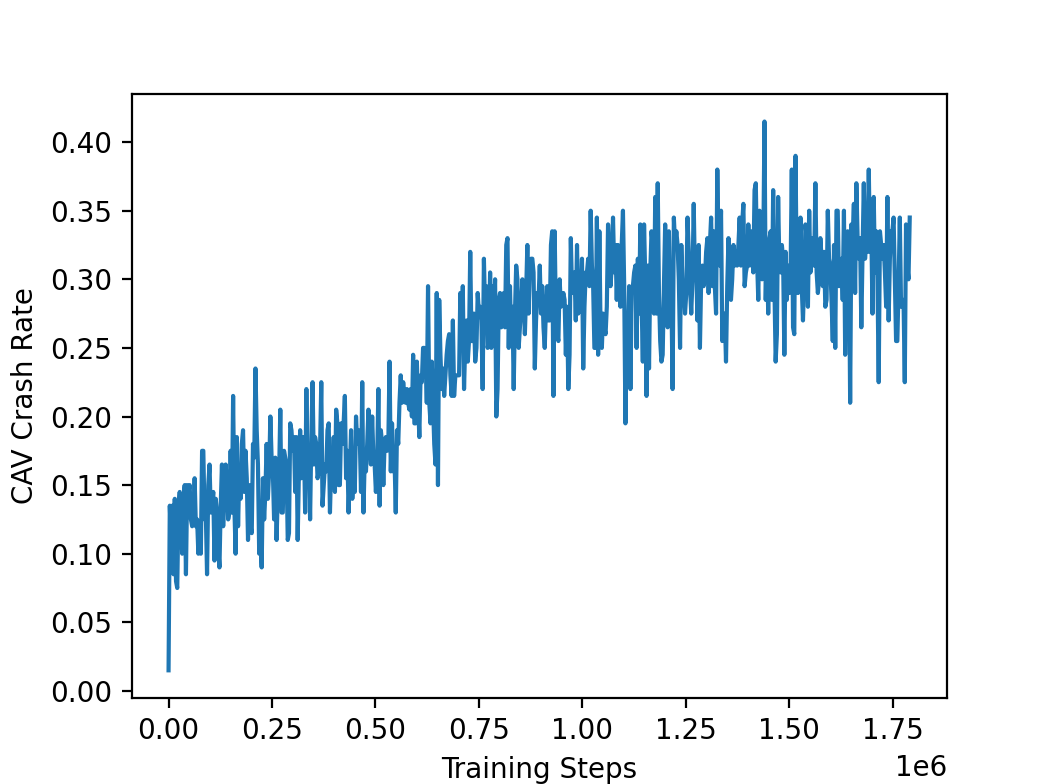}
  \caption{DQN Training Result}\label{fig:DQN_train}
\end{figure}

We implement the DQN \cite{mnih2015human} method using Pytorch and SGD as the optimizer. Detailed experiment hyper-parameters are listed in Table \ref{tab:parameter}. In the dueling neuron network architecture \cite{wang2016dueling}, there are 4 fully connected layers, each with 128 units. As is shown in Figure \ref{fig:dueling}, the dueling network splits into two streams of fully connected layers: the value stream and the advantage stream. Each stream has two fully-connected layers with 128 units. The final output layer of the state stream and the value stream are also fully connected. The value stream has 1 output and the advantage stream has 33 outputs (pre-defined discrete actions). The value stream output and the advantage stream output are combined using Equation \ref{dueling_def}. During the training process, $+1$ reward is given when BVs successfully crash into the CAV, and $-1$ reward is given when BVs crash into each other. Otherwise, we give $0$ reward. The experiments are implemented on Ubuntu 18.04 LTS with i9-9900k CPU, RTX 2080 TI GPU, and 64gb of RAM. The agent was trained for 4 days until it reaches the performance limit, and the training process of the agent can be seen in Figure \ref{fig:DQN_train}.

\begin{table}[h!]
	\caption{EXPERIMENTS HYPER-PARAMETERS}\label{tab:parameter}
	\begin{center}
		\begin{tabular}{l l l l}
			Hyper-parameter & Value \\\hline
			mini-batch size & 16 \\
			replay memory size & 1e6\\
			discount factor & 1\\
			learning rate & 1e-6\\
			initial exploration & 1\\
			final exploration & 0.1\\
			replay start size & 5000\\
			target network update & 1000\\
			
			\hline
		\end{tabular}
	\end{center}
\end{table}

In the case study, we test one commonly used CAV model, which is constructed by the IDM car-following model and MOBIL lane-change model. To better interpret the results, CARLA simulator \cite{dosovitskiy2017carla} is used to visualize corner cases. We implement the Naturalistic Driving Environment (NDE) \cite{feng2021NADE, yan2021distributionally} as our baseline, which contains the CAV model and NDD vehicles that follow the naturalistic driving model. In the NDE, the crash cases frequency is around $1\times 10^{-6}$, which is rare. In the corner case generation environment, we control the nearest background vehicle around the AV using the trained DQN. Results show the crash case frequency is about 0.286 as shown in Figure \ref{fig:DQN_train}. Therefore, in terms of corner case generation frequency, we get about $3\times 10^6$ times more corner cases compared with the NDE.

\begin{figure}[h!]
\centering
\subfigure[BBD Front Car]{
\includegraphics[width=0.46\textwidth]{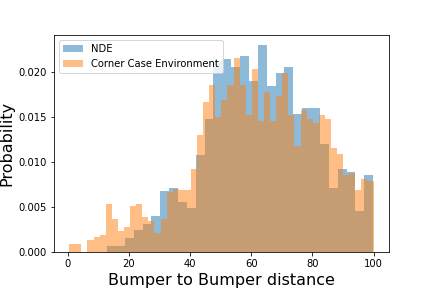}
\label{BBD_front}
}
\quad
\subfigure[BBD Rear Car]{
\includegraphics[width=0.46\textwidth]{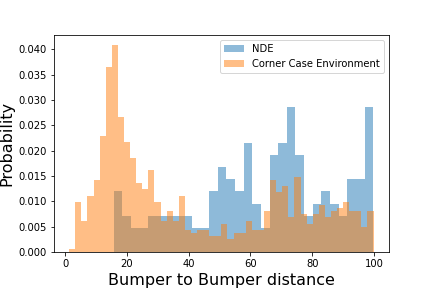}
\label{BBD_rear}
}
\quad
\subfigure[TTC Front Car]{
\includegraphics[width=0.46\textwidth]{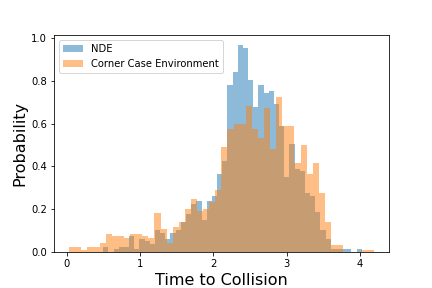}
\label{TTC_front}
}
\quad
\subfigure[TTC Rear Car]{
\includegraphics[width=0.46\textwidth]{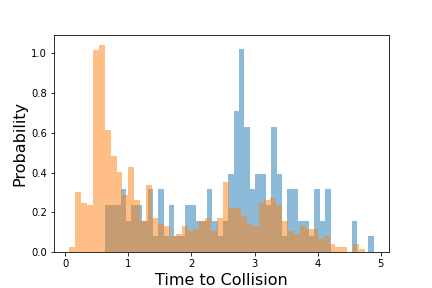}
\label{TTC_rear}
}
\caption{Comparison between NDE and Corner Case Generation Environment}
\end{figure}

To evaluate the corner case generation environment, we run about 50 miles in both NDE and the corner case generation environment respectively. The distribution of the bumper-to-bumper distance (BBD) and time to collision (TTC) metric are calculated to compare the difference between NDE and the corner case generation environment. The comparison of bumper to bumper distance can be seen in Figure \ref{BBD_front} and Figure \ref{BBD_rear}, and the comparison of TTC can be seen in Figure \ref{TTC_front} and Figure \ref{TTC_rear}. From these figures, we can see that the testing vehicle in the corner case generation environment has much smaller bumper-to-bumper distance and TTC with both the front car and the rear car. It suggests that the corner case generation environment is much riskier than the NDE.

To further test whether the corner case environment could fit non-surrogate CAV models, a CAV model based on reinforcement learning is introduced as another testing model. When the corner case environment is utilzied for the surrogate IDM-based model, we get approximate $28.6\%$ crash probability, while for the RL-based testing model we get $10\%$ crash rate.

\begin{figure}[h!]
  \centering

  \includegraphics[width=1\textwidth]{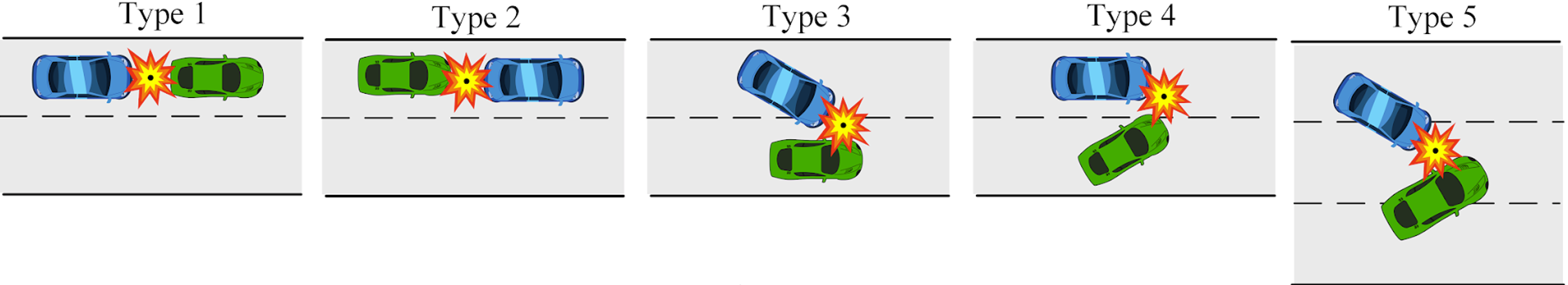}
  \caption{Illustration of Crash Types}
  \label{fig:crash_type_ill}
\end{figure}

\begin{figure}[h!]
  \centering

  \includegraphics[width=1\textwidth]{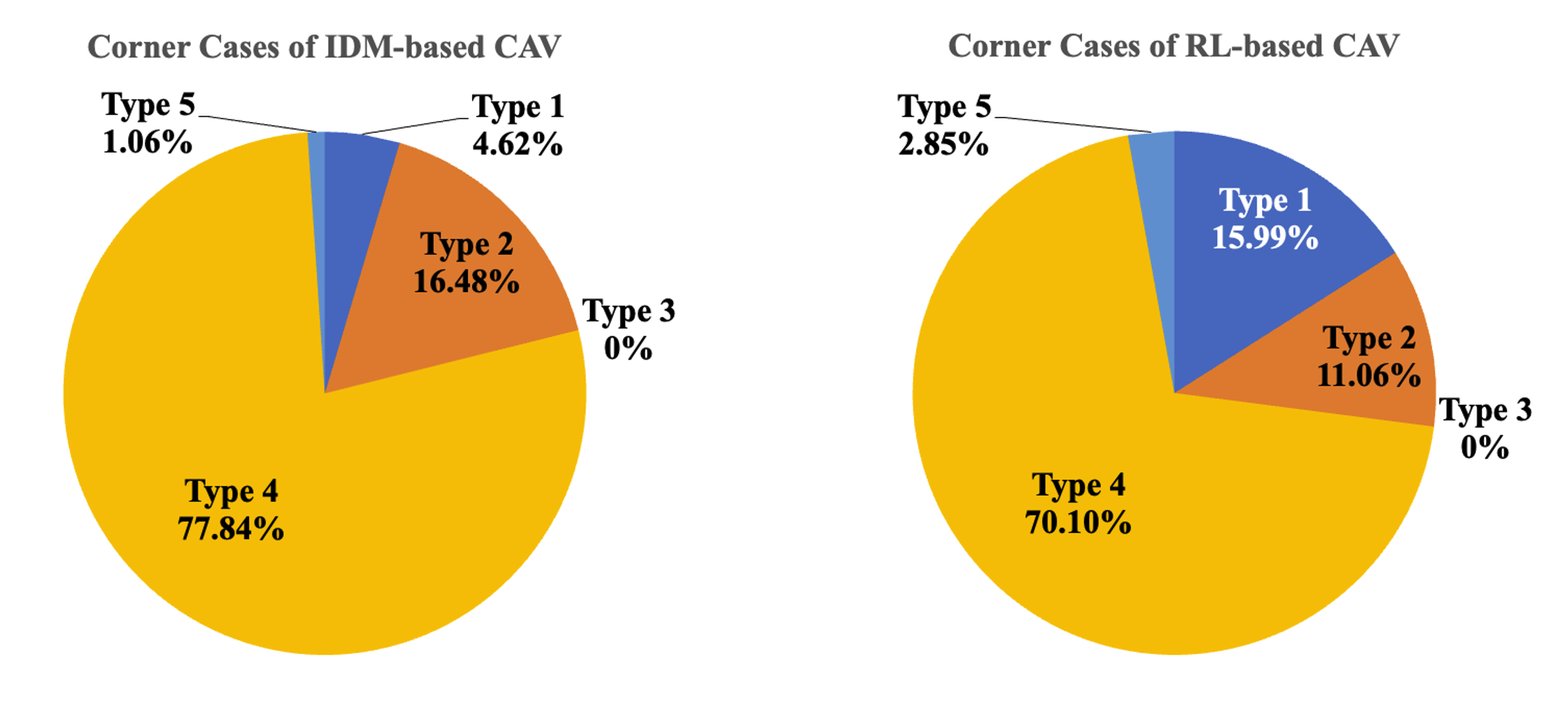}
  \caption{Distributions of Crash Types}
  \label{fig:crash_type_dis}
\end{figure}

\begin{figure}[h!]
  \centering

  \includegraphics[width=1\textwidth]{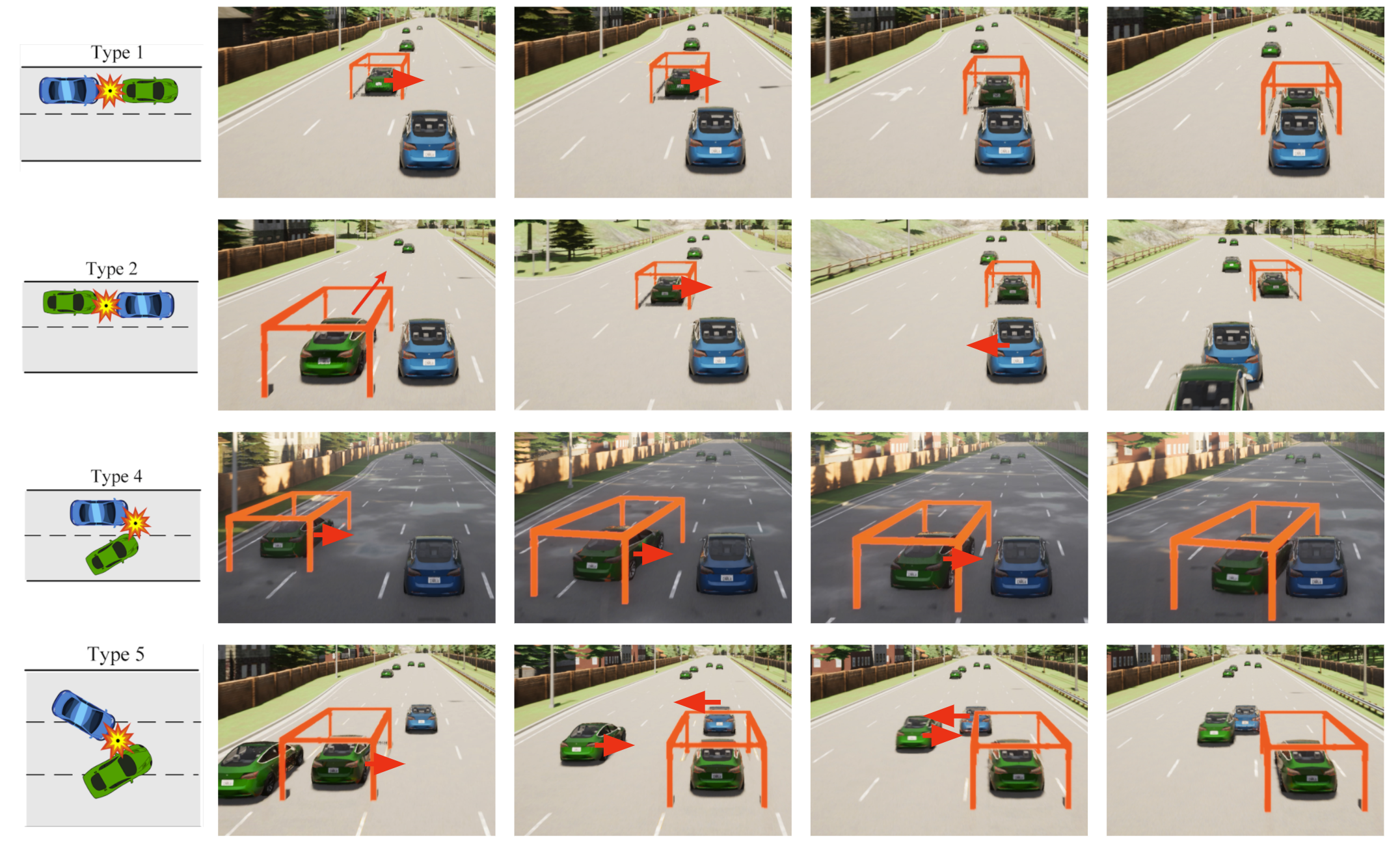}
  \caption{Demonstration of Each Crash Type in CARLA}
  \label{fig:crash_type_carla}
\end{figure}

Additionally, a logic-based crash type analysis is introduced to help illustrate the generated corner cases in detail. In this study, we adopted the crash type diagram defined by the Fatality Analysis Reporting System (FARS), which is a nationwide census provided by National Highway Traffic Safety Administration (NHTSA). Specifically, we categorize the generated corner cases into five types according to the positions and angles of the CAV and the BV involved in each crash. Detailed illustration of the categories can be seen in Figure \ref{fig:crash_type_ill}, where the blue vehicle denotes the CAV and the green vehicles denote BVs. The crash type distribution of generated corner cases of IDM-based CAV and RL-based CAV can be seen in Figure \ref{fig:crash_type_dis}. From the figure, we can see that both experiments contain the type 1, 2, 4, 5 of the overall categories, and type 4 occupies the most proportion. For each crash type, a detailed corner case demonstration is attached to illustrate the case type in detail, which can be seen in Figure \ref{fig:crash_type_carla}.

\subsection{Corner Case Analysis}
\subsubsection{Corner Case Feature Extraction}

In the generated corner cases (around 50,000 scenarios), different cases have different sizes. For the convenience of corner case analysis, we need one unified structure of selecting features from the corner cases. As discussed before, the scenario is composed of scenes, and the scene can be written as $\{x_1^{(t)}, x_2^{(t)}, \ldots, x_{m_t}^{(t)}\}$. Recall that the vehicle number $m_t$ is determined by time $t$ and is continuously changing at each time step. Therefore, to obtain features with the same shape from different time steps, we need to restrict the number of vehicles recorded. Based on domain knowledge, we select the most critical BV in the crash corner cases (i.e., the nearest vehicle in the BVs). Then we use the following equations as the extracted feature of one time step:
\begin{eqnarray}
    [r_{lon}, r_{lat}, rr, h_{CAV}, h_{BV}], \label{feature_one_time}
\end{eqnarray}
where $r_{lon}$ refers to the longitudinal relative distance, $r_{lat}$ refers to the lateral relative distance, $rr$ refers to the velocity difference, and $h_{CAV},~h_{BV}$ represent the heading angle of CAV and critical BV respectively. The feature extracted can represent the key information in a snapshot of the traffic simulation. To characterize the long-term change of the traffic state, we continuously pick features from $k$ time steps before crashes.

\begin{figure}[h!]
  \centering
  \includegraphics[width=1\textwidth]{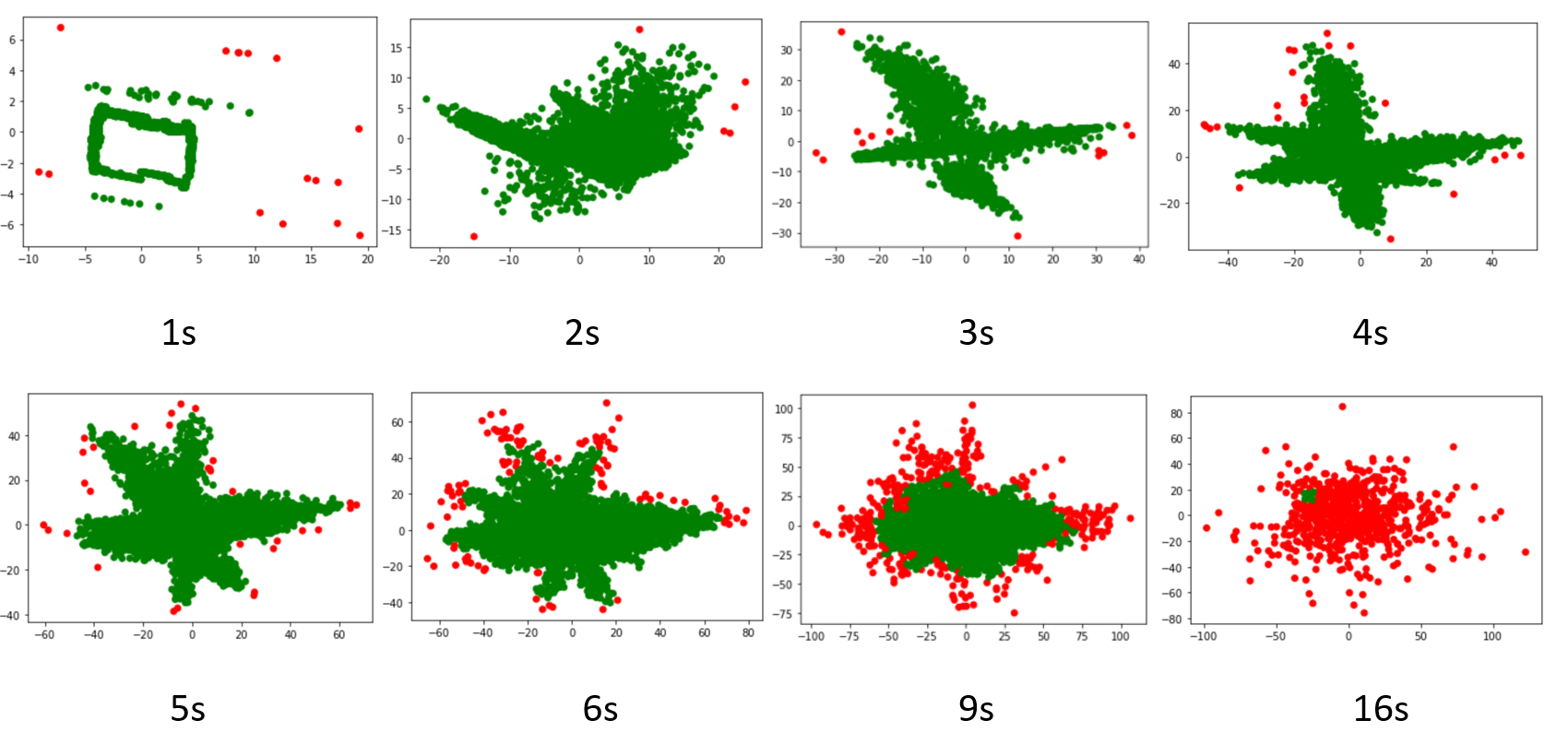}
  \caption{Corner Case Extracted Features}\label{fig:cluster}
\end{figure}

In the experiment, eight values of the time period are applied,  starting from the one time step to $15$ consecutive time steps. For each time step, 5 features are considered as shown in Equation \ref{feature_one_time}. Therefore, for $15$ time steps, 75 features are considered. To extract the crucial features from the high dimensional data, we apply the PCA algorithm and pick the first two dimensions. After the feature extraction and projection, we implement the DBSCAN clustering method to differentiate the minority and the majority of the generated corner cases. The distribution of the data points projection on the 2-D plane can be seen in Figure \ref{fig:cluster}. The data points located in the high-density area are classified as the majority (green), while the low-density ones are classified as the minority (red). From the clustering result, one obvious trend is that, with the increase of the time period, the data points seem to be less gathered. Regarding the features of $1$s or $2$s, only several data points are identified as the minority (marked as red). However, when it turns to quiet a long time period ($15$s), nearly all data points are randomly distributed over the feature space, which makes it harder to differentiate the majority and the minority. Therefore, in the following analysis, we use the $1$s data as an example to illustrate the result of the generated corner cases. It is also reasonable considering that most of the vehicle accidents in the real world involve only a small number of vehicles in a short period.

\subsubsection{Corner Case Clustering}

\begin{figure}[h!]
  \centering
  \includegraphics[width=0.6\textwidth]{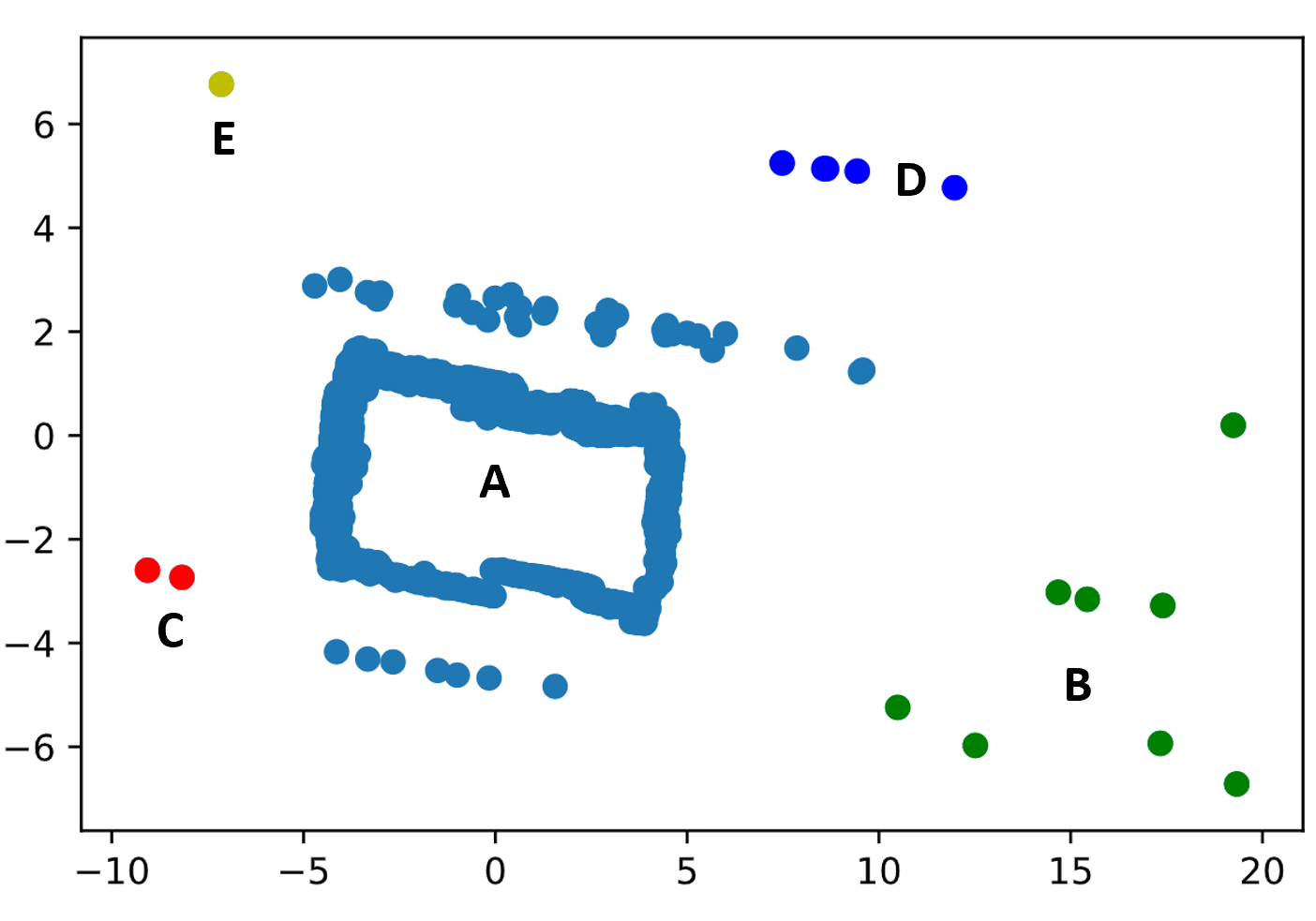}
  \caption{Corner Case Clustering Results}\label{fig:one_dim}
\end{figure}

\begin{figure}[h!]
  \centering
  \includegraphics[width=1\textwidth]{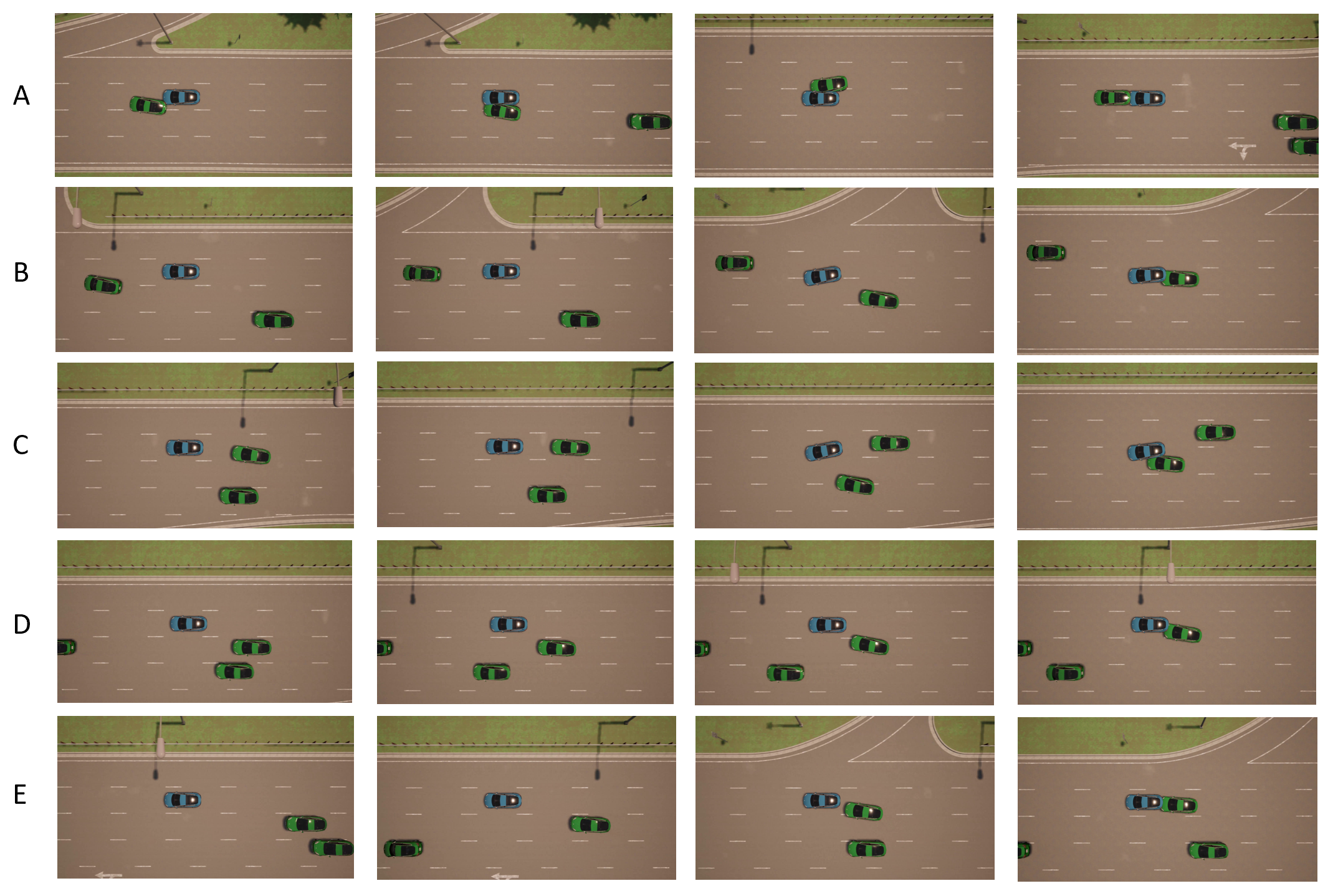}
  \caption{Corner Case Examples}\label{fig:sample}
\end{figure}

After applying the DBSCAN method on the generated corner cases, we further cluster the minority of generated corner cases. As shown in Figure \ref{fig:one_dim}, the majority (A part) are distinguished from the minority. We can see that the majority of the corner cases form a rectangle in the projected PCA feature space. In this rectangle, most cases share some characteristics in common: the CAV is running straight on the road, and the background vehicle is directly crashing into the CAV from different angles and different positions as shown in Figure \ref{fig:sample}A. In this type of crash, the crash angle and crash relative position parameters are continuous in the parameter space, while other parameters are the same. Therefore, even though it seems that there are many different kinds of crashes in this crash type, they are closely connected in the PCA feature space.

From the 2-D projection of the feature vector, we can see that the minority can be assigned to several clusters, while the majority is closely connected. Therefore, in this case, we apply the clustering method (K-means) on the minority, resulting in 4 clusters (B, C, D, E in Figure \ref{fig:one_dim}). The clustering process took $0.097s$ on a laptop with i7-8750h CPU and 24Gb RAM. For each cluster, cases of each cluster share similar internal properties. For example, eases of B cluster in Figure \ref{fig:one_dim} and Figure \ref{fig:sample} demonstrate the rear-end cases, in which one BV cuts in the CAV's lane and forces the CAV to change its lane, making the CAV crash into another BV in the original lane behind the CAV. Cases of C cluster indicate another crash type: CAV and BV change their lanes simultaneously and crash into each other. Cases of D cluster show that the BV tries to change into the lane of the CAV and causes crashes when making the lane-change decision. The only case in the E cluster behaves similarly to the cases in the D cluster. However, there are some slight differences: the crash in the E cluster happens after the BV changes into the lane of the CAV and can be defined as a rear-end crash.

\subsection{"NDD Bounded" Case Study}

\begin{figure}[h!]
  \centering
  \includegraphics[width=0.6\textwidth]{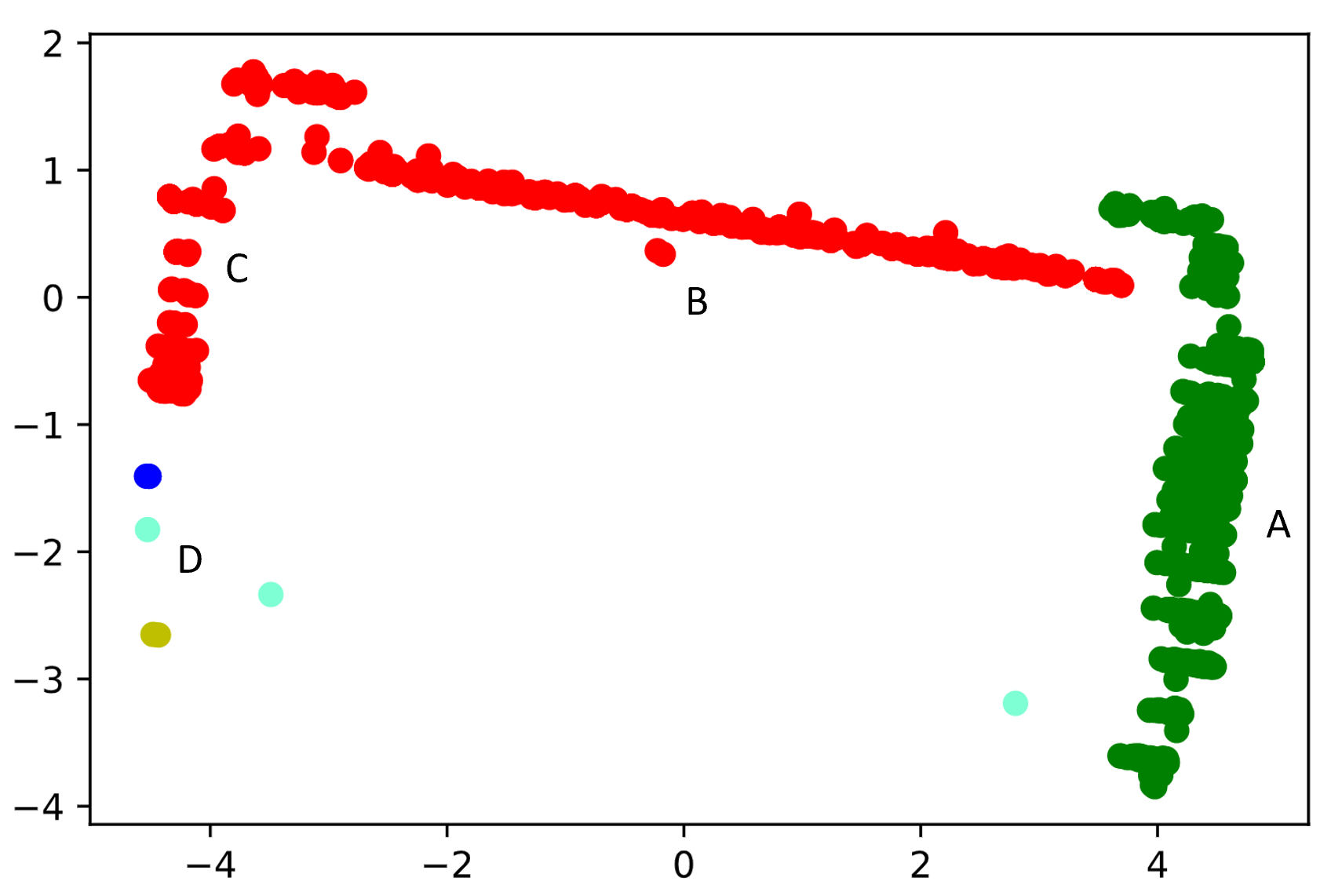}
  \caption{Corner Case Clustering Results in the "NDD Bounded" Case Study}\label{fig:NADE_cluster}
\end{figure}

We also provide the analysis results of another "NDD bounded" generation method. In Equation \ref{optimization}, the controlled BV can choose any actions in the pre-defined action space. However, in the NDE, the BVs generally have limited choices of actions. To solve this issue,  we slightly modify the optimization problem in Equation \ref{optimization} by adding action constraints. By applying the constraints, we can restrict the available actions of BVs, so the agent in the environment can only choose the actions which are possible in the NDE. The new optimization problem can be written as follows:
\begin{eqnarray}
\max_{\rho} \quad & P(A|\theta,\rho),\\
\mbox{s.t.}\quad
&\pi_{\rho}(u_k|s_k,\theta)=0~if~P(u_k|s_k,\theta)=0. \label{optimization_bound}
\end{eqnarray}
Therefore, the agent trained from the modified optimization problem can only reproduce the corner cases which are likely to happen in the NDE. In this way, the generated corner cases will be much more realistic and valuable for the CAV evaluation. Using the same analysis method of the previous result, we apply PCA on the crash event data and reduce the data into two dimensions. After that, we apply the DBSCAN algorithm on the PCA projected features to get different clusters and outliers. The experiment is applied on $1627$ data samples and costs $0.0139s$ on a laptop with i7-8750h CPU and 24Gb RAM. Detailed results can be seen in Figure \ref{fig:NADE_cluster}.

\begin{figure}[h!]
  \centering
  \includegraphics[width=1\textwidth]{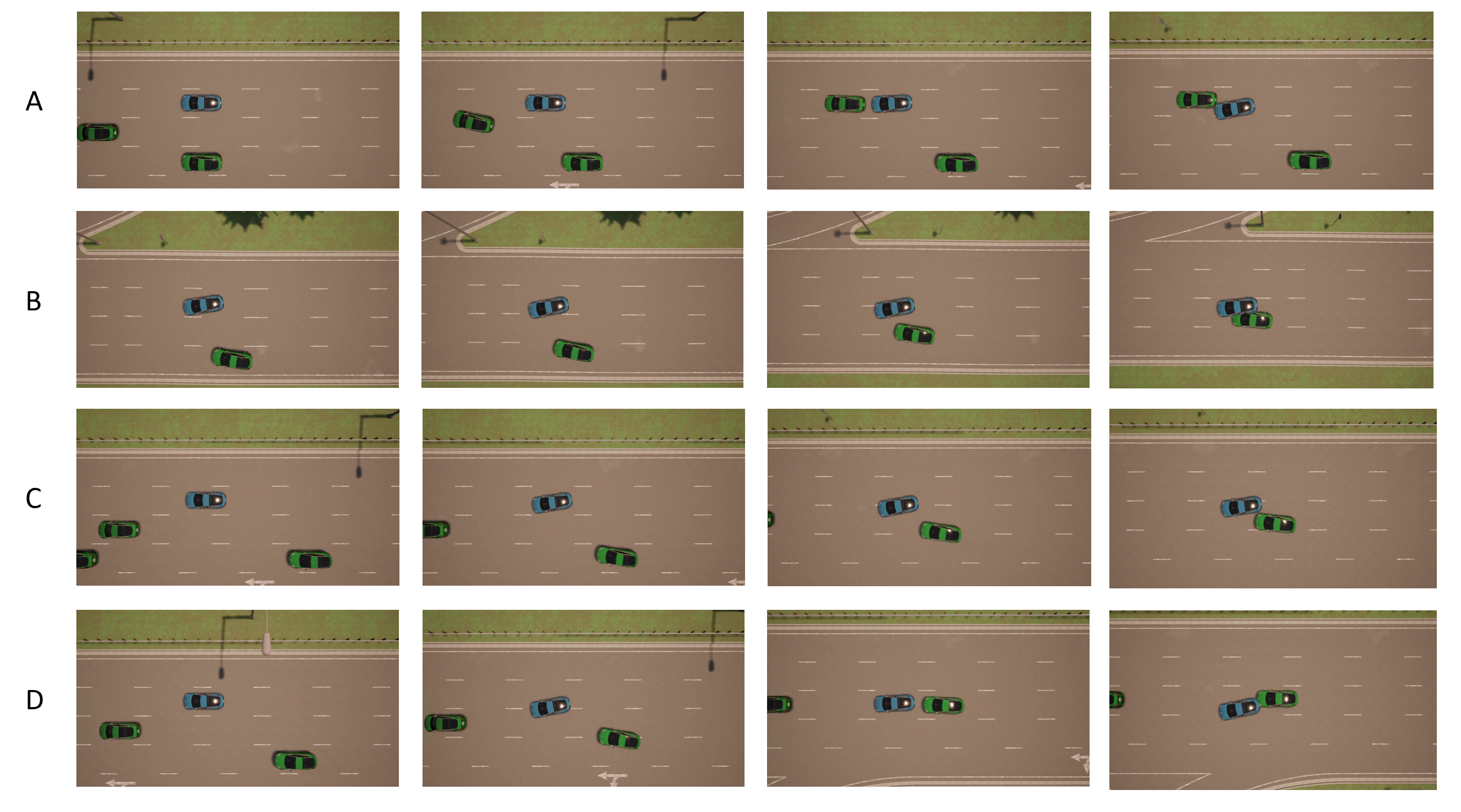}
  \caption{"NDD Bounded" Corner Case Examples}\label{fig:NADE_sample}
\end{figure}

From Figure \ref{fig:NADE_cluster}, we can see that there are two main clusters identified and several outliers (D). To analyze the results, we split one large cluster into B and C, which are deeply connected but demonstrate different data layout. As shown in Figure \ref{fig:NADE_sample}, data points in A area share similar properties: the BV suddenly cuts in the CAV and causes the rear-end collision. We define it as the "aggressive cut-in" cluster. B, C, and D are in nearly the same situation: CAV and BV change their lanes at the same time and crash in the middle lane. We define it as the "lane conflict" cases. Even though the causes of crashes are similar, these three different clusters have different crash snapshots. In the B cluster, the CAV and BV are involved in a side-by-side collision. In C cluster, CAV(BV) crashes into BV(CAV) at the rear of the car. In D cluster which is identified as the relatively rare events (minority) in the generated corner cases, we can see that the CAV and BV do not crash until they change to the same lane, after which the rear-end collision happens. Therefore, even though the last three clusters are corner cases caused by the same reason, the decision variables and conditions are different case by case. In the outlier part (D), we can get some cases with extreme decision variables.
 
The extracted valuable corner cases can provide insight for the further improvement of the CAV model. Although some cases are inevitable, there are cases caused by the flaws of the CAV model, and thus can be avoided by more intelligent CAVs. For example, for the CAV model in the case study, one significant limitation is the lack of behavioral competency for lateral collision avoidance, especially during the lane changing process, as shown in Figure \ref{fig:NADE_sample} B, C, D clusters.

\section{Conclusion}

In this paper, we propose a decision-making corner case generation and analysis method for CAV testing and evaluation purpose. By utilizing MDP formulation and DRL techniques, the corner cases of highway driving environment are purposely generated with a higher probability, comparing with the NDE. After generating the corner cases, the valuable corner cases are further identified by the corner case analysis method, including feature extraction and clustering techniques. Two case studies are provided to validate the proposed methods. Results show that the valuable corner cases can be effectively generated and identified, which are helpful for CAV evaluation and development by revealing flaws of the given CAV model. Future studies may focus on introducing more different traffic participants (pedestrians, traffic lights, etc.). Furthermore, the corner case generation for urban driving environment deserves more investigation, including intersection and roundabout scenarios.

\section{ACKNOWLEDGMENTS}

The authors would like to thank the US Department of Transportation (USDOT) Region 5 University Transportation Center: Center for Connected and Automated Transportation (CCAT) of the University of Michigan for funding the research. The authors would like to thank Miss Miaoshiqi Liu for valuable suggestions. The views presented in this paper are those of the authors alone.

\section{AUTHOR CONTRIBUTION STATEMENT}
The authors confirm contribution to the paper as follows: study concept and design: Haowei Sun, Shuo Feng, and Henry Liu; simulation platform construction: Haowei Sun, Xintao Yan; analysis and interpretation of results: Haowei Sun, Shuo Feng, and Henry Liu; draft manuscript preparation: Haowei Sun, Shuo Feng, and Henry Liu. All authors reviewed the results and approved the final version of the manuscript.
\newpage

\bibliographystyle{trb}
\bibliography{corner_case_generation}
\end{document}